\documentclass{article}
\usepackage{ijcai24}

\usepackage{times}
\usepackage{soul}
\usepackage{url}
\usepackage[hidelinks]{hyperref}
\usepackage[utf8]{inputenc}
\usepackage[small]{caption}
\usepackage{graphicx}
\usepackage{amsmath}
\usepackage{amsthm}
\usepackage{booktabs}
\usepackage{algorithm}
\usepackage{algorithmic}
\usepackage[switch]{lineno}
\usepackage{multirow}
\usepackage{subcaption}

\title{Reinforcement Learning for Graph Coloring: Understanding the Power and Limits of Non-Label Invariant Representations}

\author{
Chase Cummins
\And
Richard Veras\\
\affiliations
University of Oklahoma\\
\emails
chasecummins1@ou.edu,
richard.m.veras@ou.edu
}

\begin{document}

\maketitle

\begin{abstract}
Register allocation is one of the most important problems for modern compilers. With a practically unlimited number of user variables and a small number of CPU registers, assigning variables to registers without conflicts is a complex task. This work demonstrates the use of casting the register allocation problem as a graph coloring problem. Using technologies such as PyTorch and OpenAI Gymnasium Environments we will show that a Proximal Policy Optimization model can learn to solve the graph coloring problem. We will also show that the labeling of a graph is critical to the performance of the model by taking the matrix representation of a graph and permuting it. We then test the model's effectiveness on each of these permutations and show that it is not effective when given a relabeling of the same graph. Our main contribution lies in showing the need for label reordering invariant representations of graphs for machine learning models to achieve consistent performance.
\end{abstract}

\section{Introduction}

Today's modern compilers can optimize most general purpose code fairly quickly using heuristics that trade runtime performance for compile time. But, as we look to squeeze every last drop of performance out for compute critical codes, we can approach these problems with more aggressive compilation techniques using reinforcement learning to search for more efficient output code. In this paper we explore the intersection of reinforcement learning and compiler optimization by casting a register allocation problem as a graph coloring problem for a model to learn from.

Our contributions consist of:
\begin{itemize}
    \item Evaluating Deep Q-Network (DQN) and Proximal Policy Optimization (PPO) models for graph coloring
    \item Evaluating various reward functions' effectiveness when coloring graphs
    \item Building an expandable Gymnasium environment for graph coloring
    \item Showing the need for label reordering invariant representations of graphs
\end{itemize}

\section{Theory}
In this section, we will look at and provide brief introductions to a few important concepts to our research: graph coloring and register allocation, Q-learning, Deep Q-learning and Proximal Policy Optimization. A basic understanding of these topics is important to understanding the background and methodology that we will use as we implement, test, and evaluate our graph coloring environment with DQN and PPO machine learning models.

\subsection{Graph Coloring \& Register Allocation}
A graph is given by a set of nodes or vertices $V$ and a set of edges $E$ that each connect two nodes. Graph coloring is the process of coloring the nodes of a graph such that every node has a different color than any node it is connected to. For example, if node 1 has color blue and is connected to nodes 2, 3, and 4, then none of nodes 2, 3, and 4 may be colored blue. It is not true that nodes 2, 3, and 4 cannot be the same color, unless of course there is an edge between any of them. The minimum number of colors needed to color a graph properly is known as its \emph{chromatic number}. 

The problem sounds relatively simple, but in practice as graphs grow in size, the ability to color a graph optimally (with the least amount of colors) in a reasonable amount of time diminishes rapidly. Coloring a graph with its chromatic number of colors is NP-Hard and, in general, to color a graph with $V$ nodes with $m$ colors it takes $O(m^V)$ time, which is NP-Complete \cite{graph-color}.

A particularly important application of graph coloring is in the area of compilers. Here the critical task of register allocation -- mapping  temporary variables to a small number of machine registers -- is cast in terms of a graph coloring problem. A Register Inference Graph (RIG) is built where each variable in a block is represented as a vertex and a connection exists between two vertices if their value must be made available at overlapping points in time. The objective is then to find a valid $k$-coloring, where $k$ is the number of available machine registers. If a valid coloring is not found, then the code is modified to "spill" a temporary value to memory until it is needed later in the code, where an available register is then "filled" with the value and the process is repeated until a valid coloring is found. While introducing the "spill" and subsequent "fill" alleviates "register pressure" by decreasing the constraints in the RIG, this is done at the expense of the performance of the resulting compiled code. Because general graph coloring NP-Complete, modern compilers use heuristics that reduce compile time at the expense of efficiency.

\begin{figure*}
\begin{subfigure}[b]{0.49\textwidth}
    \centering
    \includegraphics[width=\textwidth]{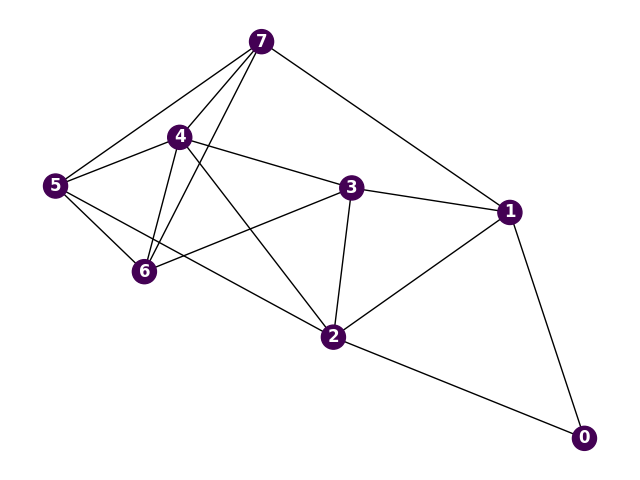}
    \caption{An uncolored graph with 8 nodes or vertices.}
    \label{fig:uncolored-graph}
\end{subfigure}%
\begin{subfigure}[b]{0.49\textwidth}
    \centering
    \includegraphics[width=\textwidth]{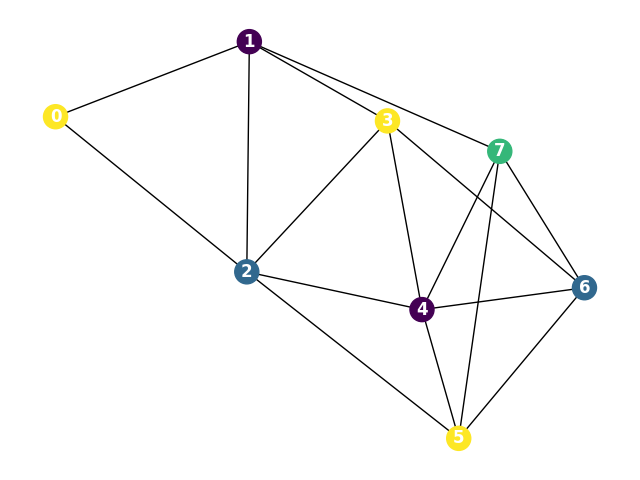}
    \caption{A proper 4-colored version of the same 8-node graph.}
    \label{fig:colored-graph}
\end{subfigure}
\end{figure*}

\subsection{Q-Learning}
One equation that is significant for some types of reinforcement learning such as Q and Deep Q-learning is the \emph{Bellman Equation} \cite{Mayank_2018} .
\begin{displaymath}
    Q^{new}(s,a) = (1-a)\ Q(s,a) + a\left(R_{t+1} + \gamma\ max\ 
    Q(s',a')\right)
\end{displaymath}
The Bellman equation is the primary means by which the agent learns, and the primary function resides in (\textit{state}, \textit{action}) pairs shown above as forms of $(s,a)$.

\begin{itemize}
    \item  $Q(s,a)$ in the equation is the \textit{Q-value} associated with the current (state, action) pair.
    \item $Q^{new}(s,a)$ is the new Q-value associated with the (state, action) pair.
    \item $a$ is the \textit{learning rate} that the programmer chooses where closer to $1$ updates the Q-value more while closer to $0$ has the Q-value update slower.
    \item $R_{t+1}$ is the reward the agent receives transitioning from state $s$ to state $s'$ by action $a$.
    \item  $\gamma$ is the discount factor where closer to $0$ means immediate rewards are given preference and closer to $1$ means future rewards are prioritized.
\end{itemize}.    

The Q-table is the most basic version of Q-learning in that it associates a specific value with each (state, action) pair and therefore there is a table of values that is stored. In Table \ref{Q-table example} we show an example of what the q-table might look like. Each $Q(s_i,a_i)$ represents some q-value for that specific (state, action) pair, which is updated to $Q^{new}(s,a)$ by the Bellman Equation explained above. The values in the table are often set randomly at the beginning of each trial to allow it to converge to an optimal solution faster. This version of Q-learning is very efficient and effective for small problems, but one can see that as the number of states and actions possible increases, the size of the Q-table increases exponentially.

\begin{table}
  \caption{An example of a Q-table.}
  \label{Q-table example}
    \begin{tabular}{cc|c|c|c|c|l}
        \cline{3-6}
        & & \multicolumn{4}{ c| }{States} \\ \cline{3-6}
        & & $s_1$ & $s_2$ & $\ldots$ & $s_n$ \\ \cline{1-6}
        \multicolumn{1}{ |c  }{\multirow{4}{*}{Actions} } &
        \multicolumn{1}{ |c| }{$a_1$} & $Q(s_1,a_1)$ & $Q(s_2,a_1)$ & $\ldots$ & $Q(s_n,a_1)$ &     \\ \cline{2-6}
        \multicolumn{1}{ |c  }{}                        &
        \multicolumn{1}{ |c| }{$a_2$} & $Q(s_1,a_2)$ & $Q(s_2,a_2)$ & $\ldots$ & $Q(s_n,a_2)$ &     \\ \cline{2-6}
        \multicolumn{1}{ |c  }{ } &
        \multicolumn{1}{ |c| }{$\ldots$} & $\ldots$ & $\ldots$ & $\ldots$ & $\ldots$ &  \\ \cline{2-6}
        \multicolumn{1}{ |c  }{}                        &
        \multicolumn{1}{ |c| }{$a_n$} & $Q(s_1,a_n)$ & $Q(s_2,a_n)$ & $\ldots$ & $Q(s_n,a_n)$ &  \\ \cline{1-6}
\end{tabular}
\end{table}

\subsubsection{Deep Q-Learning}
For anything beyond a small number of states and actions the Q-table will grow intractably large. In \cite{mnih} the authors demonstrate the use of deep convolutional neural networks to approximate the Q-table, which we call Deep Q-Learning.

Neural networks consist of node layers divided into three parts: input layer, hidden layer(s), and output layer. Beginning at the input layer, data is fed into the neural network going through each hidden layer, finally arriving at the output layer where the output of the nodes is the probability that an associated action should be taken. The node layers are made up of individual nodes with weights and a threshold. Each node in a node layer is connected to one, multiple, or all nodes in the node layer before and node layer after it (with the exception of the input and output layers that are only connected in one direction). Each incoming connection to a node is assigned a weight and the weight is the significance of the input node's value to the receiving node. The (activation) threshold is the required level of input for the node to pass data to the next layer of the network. If the level of total input is greater than the threshold, the node "fires" and sends input to its output connections. This can be understood by the below formula where $t$ is the threshold and $w_i$ is the weight for each input value $x_i$.

\begin{displaymath}
        out = \begin{cases}
      a > 0 & \text{if $\sum w_ix_i > t$}\\
      0 & \text{else}
    \end{cases}
\end{displaymath}

Deep Q-learning consists of three parts: main DQN, target DQN (optional), and replay memory. The main DQN is what we are using to approximate the Q-table. It will take in the current state to its input layer and output "Q-values" at each output layer node. Each possible action is mapped to an output layer node and therefore we have our Q-values for a given (state, action) pair just like with Q-tables \cite{Wang_2021}. 

\subsection{Proximal Policy Optimization}
Proximal Policy Optimization (PPO) \cite{PPO-paper} like DQN is an algorithm for Reinforcement Learning, however, rather than approximating the value function (Q-table in DQN), it learns the probability that an action in a given state will lead to the maximum reward. These probabilities are captured in a "policy" that is updated according to following formula:

$$L^{CLIP}(\theta) = \bar{E}_t\left[min\left(r_t(\theta)*\bar{A}_t, clip\left(r_t(\theta),1 - \epsilon,1 + \epsilon\right)*\bar{A}_t\right)\right]$$
$$r_t(\theta) = \frac{\pi_{\theta}(a_t|s_t)}{\pi_{\theta_k}(a_t|s_t)}$$

\begin{itemize}
    \item $r_t(\theta)$ is the probability ratio between the new and old policy. If the ratio is greater than 1, then the action is more probable for the current policy than the previous. If it less than 1, the opposite is true.
    \item $\bar{A}_t$ is the advantage, which simply means whether the action was good or bad, with positive being good and negative bad.
    \item $\epsilon$ is a hyper-parameter that controls how much the new policy can change from the old policy.
\end{itemize} 

In the formula, the $clip$ function is used to control how far the policy updates, and we take the $min$ because we want to make small positive updates, but allow larger negative updates if we have made a bad action more probable. The formula is somewhat complex, but more understanding can be found in the original paper \cite{PPO-paper} and online resources.

PPO's main advantage lies in its clipping function because it allows for more stability when training. With every update, the policy only moves a small amount and therefore with PPO we can train multiple times on the same sample. This allows us to squeeze all of the information we can from a sample, which is only possible because of the conservative policy updates \cite{PPO-paper}.

\section{Mechanism}
In understanding graph coloring and its relation to register allocation, Deep Q-learning, and PPO, we can implement the graph coloring environment. This process includes creating the important reward functions that govern the model's learning ability.

\subsection{Gymnasium Environment}
Machine learning frameworks rely on some kind of environment for the model to observe, take actions in, and get reward from to learn. We have chosen to use an OpenAI Gymnasium environment since it is commonly used and eliminates a lot of the extra hassle of connecting an environment to a machine learning model. The environment is crucial to the model's ability to learn as it defines, what it can see, the actions it can take, and how good those actions are. 

We created an environment called \verb|GraphColoring| that captures the mechanisms of a graph coloring problem. The environment stores a graph as a simple adjacency matrix along with a list of numbers that correspond to the color of each node of the graph. We initialize all the nodes to '0' which we define as uncolored. We set the max number of color options to the maximum number of nodes, but this could be easily changed to narrow the size of the action space for large graphs or be set to the number of registers. The step function changes the color of a node by changing the number in its index in the colors list and then computes the reward for the action which also checks if the graph is colored properly. The environment is reset by setting all the colors to '0' in the color list.

\subsubsection{Permutation Function}
To later test the model's viability for permutations of the same graph we created a permutation function in the environment that takes the initial graph and randomly reassigns the order of the vertices in the adjacency matrix. For a graph of $N$ nodes, there are $N!$ ways to represent it in as an adjacency matrix.

\subsection{Initial Reward Functions}
Every environment must have a \verb|step()| function that takes an action and returns the reward for taking that action. The reward function is perhaps the most important part of the environment because every model is trying to maximize the reward it receives. We started our experimentation with a simple DQN we built with one hidden layer.

On the first implementation of the reward function, we simply iterated through every edge and checked if the nodes connected by the edge were properly colored. If they were, we added some positive reward, and if not we added a negative reward. If either of the nodes were uncolored, then we left the reward as is. Every non-zero reward was divided by what we called the \verb|color_factor|, which was simply the number of colors used or 1 if the graph was uncolored. These rewards accumulated as each edge was iterated over and the goal was to simply have more reward as more edges had properly colored nodes. This would then be scaled so that the less colors that were used, the greater the reward would be. The results were poor and the model did not appear to be learning. We experimented with having the reward increase exponentially as less colors were used and whether we scaled the reward with the number of colors used at all, but the output still appeared random. It was after this that we decided to try PPO instead of DQN for our model. 

\subsection{Final Reward Function}
For the PPO model, we used stable-baselines3's implementation of PPO which allowed us quick setup and experimentation with just a few lines of code \cite{stable-baselines3}. The only customization we did to the model was adding a linear learning rate scheduler to decrease the learning rate linearly over the training time steps. We found that this helped decrease variance as the model approached optimal reward. After switching to the PPO model, it became obvious that the model was actually learning to maximize the reward. We continued with the same reward function, but realized quickly that the model was smarter than we were when creating the function. It learned to exploit "neutral" moves and the fact that the positive rewards weighed heavier than the negative. Instead of learning to color the graph in as few moves and colors as possible, it was taking hundreds of steps and gaining reward in massive quantities. We realized that it would be better to evaluate the reward by comparing the current state to the previous state instead of just evaluating the current state. This meant discovering what action the model took and determining if it was a good, bad, or neutral action. 

This took some tinkering, because the model would exploit any "neutral" moves that could give it an advantage to get more reward in the end. After some trial and error, we found a reward function that the model could not exploit and that effectively led it to minimize the number of steps and colors used. In essence, the function adds 5 reward for correctly coloring a node, subtracts 5 reward for taking off a correct coloring, and subtracts 10 reward if it incorrectly colors a node. All other actions are neutral with 0 reward, but the reward starts at -1 to punish the model for taking unnecessary moves. This reward is then divided by the \verb|color_factor| mentioned previously to incentivize the model to use less colors. This reward function was also tried on the DQN after implementation for the PPO model.

The environment can use any graph represented as an adjacency matrix or use the NetworkX library to read in an adjacency list from a file and convert it. The environment is expandable to graphs of any size, but the largest we tested was of 16 nodes giving us an adjacency matrix of 16x16.

\section{Evaluation \& Analysis}
Now, understanding the necessary theory of graph coloring as well as the mechanisms of our Gymnasium environment and machine learning models, we can look at the results of both the DQN and PPO models. From these results, we will form our conclusions on the effectiveness of reinforcement learning algorithms for graph coloring using non-label reordering invariant representations of the graphs.

\subsection{DQN Results}

\begin{figure}
    \centering
    \includegraphics[width=\linewidth]{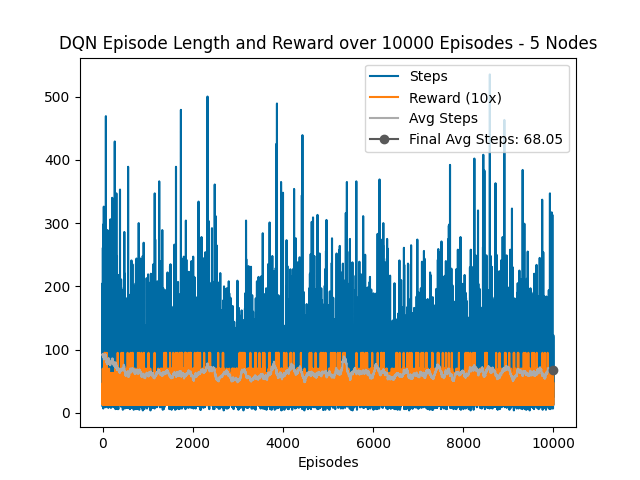}
    \caption{A plot showing the training of the DQN over 10,000 episodes given a graph of 5 nodes. The average number of steps needed is close to uniform over the training period, neglecting an initial drop at the start, showing that the DQN was failing to improve past the 68 step average it finished at.}
    \label{fig:dqn}
\end{figure}
The initial reward functions created on the DQN did not contribute to learning, although we are not sure that the reward functions were entirely the problem. As shown in Figure \ref{fig:dqn} , which is using the newest reward function that was successful on PPO, the DQN still failed to learn. The reward looks evenly distributed over the 10,000 episodes and the gray line showing the average steps goes down initially, but settles around 68 steps for the remainder of the training period with no sign of improvement. This leads us to believe that the poor reward functions were not the entire problem, but instead DQN might not be able to handle a task as complex as graph coloring. This lends more support to the idea that Q-Learning is great for simple problems, but is not as effective for more complex sets of problems such as graph coloring.

\subsection{PPO Results}

\begin{figure}
    \centering
    \includegraphics[width=\linewidth]{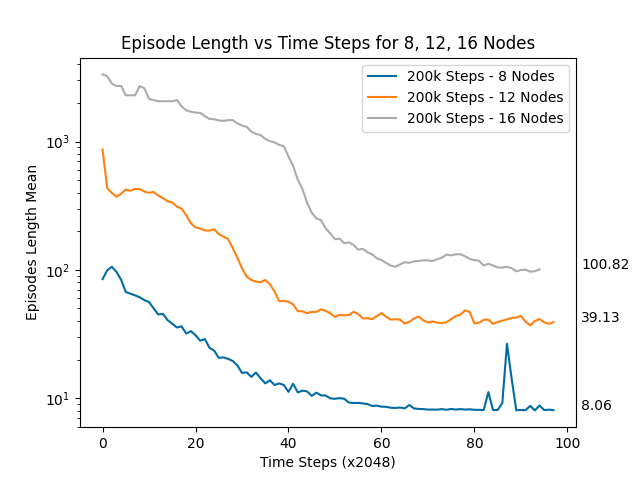}
    \caption{A plot showing the number of steps required to solve graphs of 8, 12, and 16 nodes over 200k training time steps. The model learns to solve the 8-node graph in 8 steps, the minimum number required, but fails to reach that same metric on graphs of 12 and 16 nodes over 200k time steps. Additional training was performed on the 16 node graph up to 400k time steps leading to a 50 step average in place of the 100 step average with 200k time steps. This points to a possible quadratic relationship between graph size and necessary training time.}
    \label{fig:length-with-diff-num-nodes}
\end{figure}
The PPO algorithm was much more promising and successfully learned to color an 8 node graph in 8 steps using the minimal number of colors as shown in Figure \ref{fig:length-with-diff-num-nodes}. The model was trained for approximately 200,000 time steps, a batch size of 128, and a linear learning rate scheduler with initial value \verb|3e-4|. Also shown in the figure are the results of two graphs with 12 and 16 nodes. It should be noted that the y-axis is on a logarithmic scale and that the final episode lengths (steps taken to solve) for the 12 and 16 node graphs are both significantly higher than their number of nodes, which would be the optimal number of steps.

We continued training for the 16 node graph out to 400k time steps and it reached an average episode length of about 50 steps, which was half the steps needed at 200k time steps. This leads us to believe that as graphs grow linearly in size, the time needed to train will most likely increase in polynomial or even exponential time to reach similar levels of optimization.

\begin{figure}
    \centering
    \includegraphics[width=\linewidth]{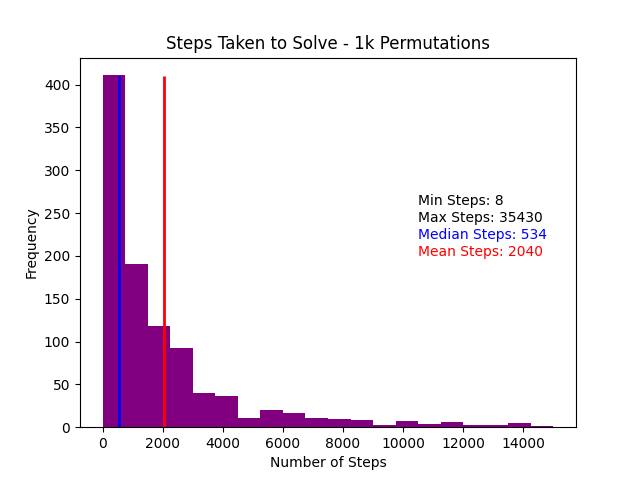}
    \caption{A histogram of steps taken to solve 1000 permutations of a graph with a trained model. The model averaged 2,040 steps and had a median number of steps needed of 534 showing that a model trained on one labeling of a graph is not well equipped to solve the same graph relabeled.}
    \label{fig:hist-orig}
\end{figure}

Shown in Figure \ref{fig:hist-orig} is a histogram showing the steps taken to solve 1,000 permutations of the original graph with a model trained to perform optimally on the original permutation. With an average number of steps at 2,040 and a median of 534, we can say confidently that the model does not often recognize that this is the same graph. The highest max was up to 35,000 steps which is about half the number of steps to try every possible combination with 4 colors ($4^8 = 65,536$). Suffice to say, the model is often ignorant of how to solve a permutation of the same graph.

\begin{figure}
    \centering
    \includegraphics[width=\linewidth]{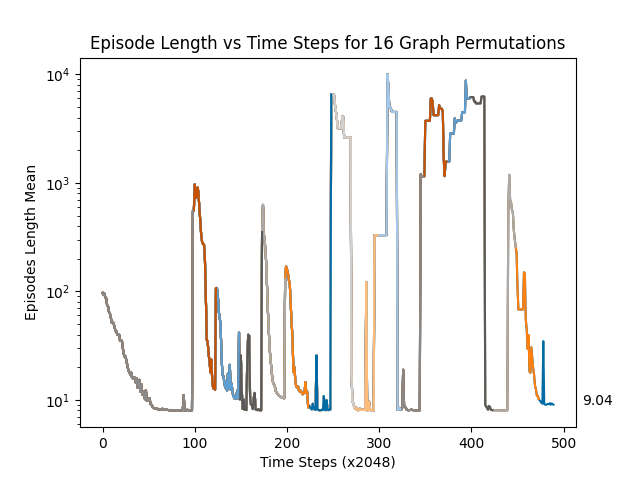}
    \caption{Plot showing the number of steps to solve over training time steps for an initial training of 200k steps on the graph and then training of 50k steps for 16 permutations of the graph. The different permutations are differently colored. As shown, the model often has to relearn how to solve the graph and sometimes it performs poorly for the entire duration of training for a given permutation.}
    \label{fig:perm-vs-ep-length}
\end{figure}

We didn't want to stop there because we reasoned that the model just needed more training on various permutations to be more effective. From this, we trained the model on the 8 node graph in the same way again with 200,000 time steps to cause it to solve optimally, but then we trained the model with 16 permutations of the graph for 50,000 time steps on each permutation. The results are shown in Figure \ref{fig:perm-vs-ep-length}. Each time the adjacency matrix is permuted, the plot changes color for that section.

As you can see, when a permutation is introduced the model practically starts over for many of the permutations. Sometimes it learns quickly how to return to near optimal solutions, and other times it seems to be clueless for the entirety of its training period. What matters most, however, is how the new model now performs when given a variety of permutations when compared with the previous model not trained on the permutations.

\begin{figure}
    \centering
    \includegraphics[width=\linewidth]{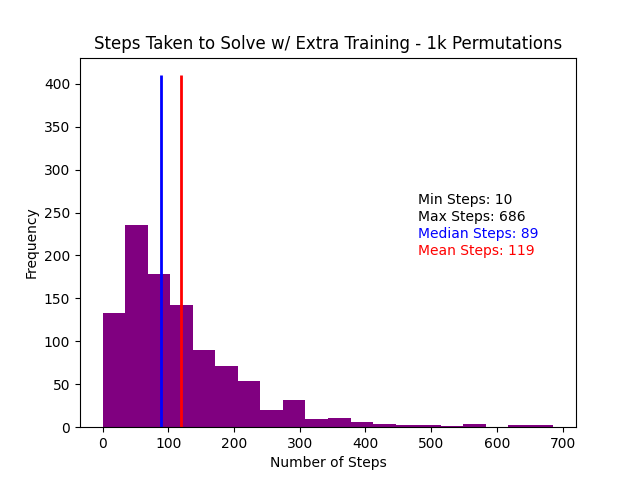}
    \caption{A histogram of steps taken to solve 1000 permutations with additional training on 16 relabelings of 50k steps each. The performance improved significantly with the additional training decreasing the median and mean steps to 89 and 119 respectively. The minimum steps increased to 10, though, implying that as the model gets better generally, it's performance on previously trained permutations will most likely decrease.}
    \label{fig:hist-new}
\end{figure}

As shown in Figure \ref{fig:hist-new}, the model does perform better now. With a mean and median number of steps needed of 119 and 89 respectively, this new model far surpasses the previous model over the 1,000 permutations tested. One unexpected observation is that the minimum is now 10 steps, even though the first permutation is the original un-permuted graph that it was originally trained to perform optimally on. It appears that the additional training actually made the model lose its ability to solve the graph optimally. While the mean and median number of steps did decrease significantly, we hypothesize that additional training would lead the model to find "average" solutions for every permutation while solving none of them optimally. However, further research is required to verify this hypothesis.

\section{Summary}
In this paper, we examined the feasibility of a non-label reordering invariant graph representation when using machine learning to solve graph coloring problems. These problems are important because register allocation, a crucial compiler optimization, can be cast as a graph coloring problem. Through experimentation, we show that DQN and PPO machine learning models fail to perform well when given relabelings of the same graph after the models have been trained to a high level of performance. From this, we conclude that in order to achieve consistent performance on graph coloring problems with machine learning models, the graph must be represented in a label reordering invariant representation.

\subsection{Related Work}

The use of reinforcement learning for combinatorial optimization is well studied and surveyed in \cite{mazyavkina2020reinforcement}. For graph coloring in particular, in \cite{huang2019coloring} the authors extend AlphaGo Zero \cite{silver2017mastering} for graph coloring on large graphs. Our work differs in that they focus on a model-based method for reinforcement learning, while our focus is on model-free methods. Additionally, the authors in \cite{das2019deep} a Long Short Term Memory (LSTM) based approach for graph coloring that is augmented with heuristics to assist in satisfying the coloring constraints.

\subsection{Future Work}
One research area that is rising to prominence and could be of interest for graph coloring is that of Graph Neural Networks (GNNs). GNNs use neural networks to transform a graph into a relabeling invariant representation without losing any attributes of the graph \cite{GNN-article}. \cite{GNN-coloring} has already shown that GNNs are a viable option for use with graph-coloring problems.

\appendix

\section*{Acknowledgments}

We would like to thank Brandon Morgan, Ph.D. candidate at the University of Oklahoma, for helping us understand how to make the DQN model work with our environment and for introducing PPO as a viable alternative to our original DQN model.

\bibliographystyle{named}
\bibliography{Paper}

\end{document}